\documentclass[12pt]{article}

\title{Testing the Reliability of ChatGPT for Text Annotation and Classification: A Cautionary Remark}
\author{Michael V. Reiss}

\usepackage{lmodern}
\usepackage[T1]{fontenc}
\usepackage{geometry}
\usepackage{setspace}
\usepackage{tikz}
\usepackage[hang,flushmargin,bottom]{footmisc}
\usepackage{scalefnt}
\usepackage{tikz}
\usetikzlibrary{decorations.pathreplacing,calc}
\usepackage{caption}

\usepackage[american]{babel}

\usepackage{csquotes}

\usepackage[style=apa, backend=biber]{biblatex}
\DeclareLanguageMapping{english}{english-apa}
\usepackage{hyperref}
\usepackage[hyphenbreaks]{breakurl}

\hypersetup{
  breaklinks = true,
  colorlinks,
  linkcolor=black,
  urlcolor=blue,
  pdfusetitle=true,
  citecolor=black
}

\usepackage{xpatch}
\AtBeginBibliography{\hypersetup{urlcolor=black}}

\usepackage{parskip}
\usepackage{microtype}

\renewcommand{\thefootnote}{\fnsymbol{footnote}}

\usepackage{titlesec}

\titleformat{\section} 
  {\normalfont\bfseries} 
  {\thesection} 
  {1em} 
  {} 

\titleformat{\subsection} 
  {\normalfont\bfseries\itshape} 
  {\thesubsection} 
  {1em} 
  {} 

\titlespacing*{\section}
  {0pt} 
  {0.85em} 
  {0.85em} 

\titlespacing*{\subsection}
  {0pt} 
  {0.5em} 
  {0.5em} 

\begin{document}
\vspace*{1cm}
\begin{center} 
\begingroup
\setlength{\baselineskip}{1.4\baselineskip}
{\scalefont{1.4} \selectfont \textbf{Testing the Reliability of ChatGPT for Text Annotation and Classification: A Cautionary Remark}}\\
\endgroup
\vspace*{1.15cm}
Michael V. Reiss\footnote{Department of Communication and Media Research, University of Zurich, Andreasstrasse 15, 8050 Zurich, Switzerland, \href{mailto:m.reiss@ikmz.uzh.ch}{m.reiss@ikmz.uzh.ch.}}\\
April 05, 2023\\

\vspace*{1.65cm}
\textbf{Abstract}\\
\vspace{0.7cm}

\begin{minipage}{0.88\textwidth}
Recent studies have demonstrated promising potential of ChatGPT for various text annotation and classification tasks. However, ChatGPT is non-deterministic which means that, as with human coders, identical input can lead to different outputs. Given this, it seems appropriate to test the reliability of ChatGPT. Therefore, this study investigates the consistency of ChatGPT’s zero-shot capabilities for text annotation and classification, focusing on different model parameters, prompt variations, and repetitions of identical inputs. Based on the real-world classification task of differentiating website texts into news and not news, results show that consistency in ChatGPT’s classification output can fall short of scientific thresholds for reliability. For example, even minor wording alterations in prompts or repeating the identical input can lead to varying outputs. Although pooling outputs from multiple repetitions can improve reliability, this study advises caution when using ChatGPT for zero-shot text annotation and underscores the need for thorough validation, such as comparison against human-annotated data. The unsupervised application of ChatGPT for text annotation and classification is not recommended.
\end{minipage}
\end{center}

\clearpage

\renewcommand{\thefootnote}{\arabic{footnote}} 
\setcounter{footnote}{0}

\section*{Introduction}

\vspace{0.5cm}
\noindent\hspace{1.5cm}%
\begin{minipage}{\dimexpr\linewidth-1.5cm}
\footnotesize
“Computer coding is 100 percent reliable, while levels of intercoder reliability among experts, and even the intracoder reliability of the same expert coding the same text at different times, can leave a lot to be desired. [...] Thus, while mechanically analysing words out of context may on the face of things seem to have an obvious cost in terms of the validity of data generated, this is off-set by a very significant gain in their reliability.” \hfill\parencite[626]{laver_estimating_2000}\\
\end{minipage}

For quite some time, automated methods assist researchers in analyzing textual data \parencite{grimmer_text_2013}, expanding content analysis to corpora sizes beyond what was possible with manual content analysis. The emergence of Large Language Models has further expanded the sophistication and fields of application of automated methods. The latest evolution of this development is ChatGPT, released in November 2022, which attracted massive public and academic attention. For research in the context of text-as-data and natural language processing, a focus is on exploring the capabilities of ChatGPT for text annotation and classification. For more traditional machine learning-based text classification, manually annotated texts were required to train machine learning classifiers. ChatGPT raises hopes to take over this labour, cost, and time intensive work by either annotating text as a basis for training data in further analyses (e.g., using established procedures that are cheaper and faster) or by directly classifying texts. Several studies already discuss and demonstrate promising zero-shot applications of ChatGPT (i.e., classification of unseen data without any training on the classification task), for example for detecting hate speech, misinformation, rating the credibility of news outlets, and more \parencite{hoes_using_2023, huang_is_2023, kuzman_chatgpt_2023, qin_is_2023, yang_large_2023, zhong_can_2023}. Even compared to manual annotation, \textcite{huang_is_2023} identify the "great potential of ChatGPT as a data annotation tool” and \textcite{gilardi_chatgpt_2023} show that ChatGPT outperforms human coders for several annotation tasks.

What has been investigated less so far is the consistency of ChatGPT’s zero-shot text classification and text annotation capabilities. While this is not necessarily an important issue when robust validation takes place (e.g., against human-annotated data), it can be an issue if ChatGPT comes to be seen as the end of manual data annotation \parencite{kuzman_chatgpt_2023} and validation efforts are neglected. The inclusion of ChatGPT in quantitative and qualitative text annotation and analysis software that offers “full-automatic data coding” \parencite{atlasti_ai_2023} justifies the concerns of unvalidated text annotation.

Alongside validity, reliability in measures is important for unbiased conclusions \parencite{ruggeri_events_2011}. However, ChatGPT is non-deterministic, which means, that identical input can lead to different outputs. Certain model parameters of ChatGPT (e.g., temperature, which steers the randomness of the output \parencite{openaia}) or the input prompt can influence this inconsistency. And while these factors can be optimized and controlled by validation, inconsistencies due to the general randomness of ChatGPT can still impair reliability of ChatGPT’s classification output or text annotations. The black-box character of ChatGPT’s inner workings add to this randomness. Therefore, this study has the aim to investigate the zero-shot consistency of ChatGPT for text classification. The study uses data of a real-world classification problem (i.e., categorizing website texts into news and not news) and investigates the influence of different temperature settings and variations in prompt instructions on the consistency of the classification output. Furthermore, the consistency across repetitions of the exact same input is assessed. By investigating the reliability and consistency of ChatGPT, this contributes to a quickly evolving body of research on ChatGPT’s capabilities for text classification and annotation but adds a more cautious perspective. The findings show that the consistency of classification outputs by ChatGPT can be below recommended thresholds, questioning the reliability of the classification and annotation results. This illustrates that validation of reliability is important, for example by comparing ChatGPT outputs with human-annotated reference data. This study only investigates reliability but the recommendation to validate outputs also pertain to validity as researchers can only be sure to measure what they intend to measure if this measure is validated.

\section*{Methods}

\subsection*{Data Collection}

This analysis is based on the classification of websites into News or not News. The websites were collected for a previous study and a subset manually annotated \parencite{reiss_dissecting_2023}. Of these, 234 website texts were randomly selected to form the basis for this investigation. The website texts (all German-speaking) were obtained for the previous study by parsing the respective html to plain text. To inform ChatGPT on the classification task it has to perform, ten different instructions were created. The first instruction is adapted from the original codebook that was used by human coders to classify the websites for the previous study. As the manual coding was done on the website screenshots, the instructions for ChatGPT were adapted as closely as possible to fit the new mode. Instructions two to ten are much shorter and less detailed but convey the same basic understanding of the classification task. All combinations of instructions and website texts form the prompts that were fed into ChatGPT. For instance, instruction two and the excerpt of one website text looked like this:

\vspace{0.4cm}
\noindent\hspace{1.5cm}%
\begin{minipage}{\dimexpr\linewidth-1.5cm}
\footnotesize
“Please classify the following text on a scale ranging from 0 to 1, where 1 means the text is news and 0 means the text is not news. Only provide a classification on the scale from 0 to 1, do not provide any explanation: Hauptnavigation Rubriken Menschen Familie Wohnen Reisen Wohlbefinden Wettbewerbe Migros-Welt Startseite Suche Newsletter abonnieren Migros-Magazin online lesen FAQ / Hilfe Artikel Tags \& Kategoriens 1. Essen, 2. Freizeit, 3. Garten 5 Ein echter Wintergarten Viele Hobbygärtner überlassen ihr Fleckchen Erde jetzt sich selbst. Nicht so Cathrin Michael: Ihr Garten ist im Winter ein fast genauso beliebter Treffpunkt wie im Sommer. […]”.\\
\end{minipage}

Furthermore, to assess the influence of different temperature settings, each prompt is tested for a low and a high temperature setting (0.25 and 1). Finally, to investigate within-consistency, all input configurations were repeated 10 times. This results in 46,800 inputs in total. Figure \ref{fig1} informs about the setup. To feed this input into ChatGPT, OpenAI’s official API and the model gpt-3.5-turbo (as of 01. April 2023, using Python 3.9) were used. The instructions, all code and data to reproduce this study’s result can be accessed via \href{http://doi.org/10.17605/OSF.IO/PRZEF}{http://doi.org/10.17605/OSF.IO/PRZEF}.

\vspace{0.4cm}
\begin{figure}[h]
\centering
\caption{Setup}
  \captionsetup{skip=10pt} 
\vspace{0.2cm}
{\sffamily

\begin{tikzpicture}[font=\sffamily]
  \foreach \i/\labeltext/\numerator in {1/Temperature/2, 2/Instructions/10, 3/Texts/234, 4/Repetitions/10} {
    \draw[thin] ({(\i-1)*3.5},0) rectangle ++(2.5,1.5) node[midway] {\shortstack{\labeltext\\\textnormal{n = }\numerator}};
  }
  
  \foreach \i in {1,2,3} {
    \draw ({(\i*3.5)-0.5},0.6) node[anchor=base,scale=1.15,align=center] {X};
  }
  
  \draw[decorate,decoration={brace,amplitude=4pt,mirror,raise=1mm}] (0,-0.1) -- (9.5,-0.1) node[midway,yshift=-0.6cm] {4680 different \textit{input configurations}};
  \draw[decorate,decoration={brace,amplitude=4pt,raise=1mm}] (3.5,1.6) -- (9.5,1.6) node[midway,yshift=0.6cm] {2340 different \textit{prompts}};
\end{tikzpicture}
\label{fig1}
}
\end{figure}

\subsection*{Analysis}

To analyze how different factors such as temperature setting, variation in instructions and repetitions of the same input configurations affect the consistency of ChatGPT’s classification outputs, the different combinations are compared. For example, to assess the impact of different temperature settings, the classification output for input configurations with a low temperature setting and a high temperature setting are compared. 

Krippendorff’s Alpha is used as a metric to assess the consistency. An Alpha of 1 indicates perfect agreement and implies, in the context of comparing temperature settings, every single text was classified in the same way when temperature parameter was high as when temperature parameter was low. Consistency for variations in instructions are analyzed likewise. Comparing the output of repetitions of the identical input therefore resembles evaluating the intra-coder reliability of ChatGPT, comparable to a scenario for a human coder.\footnote{This is not exactly true as ChatGPT has no context knowledge and memory of past interactions, since the environment is reset after each text classification. In comparison, humans learn from past texts they annotated.}  Again, Krippendorff’s Alpha is used as metric to evaluate whether the same input leads to consistent outputs. Consistencies with Krippendorff’s Alpha above 0.8 are considered reliable \parencite{krippendorff_content_2004} and should also be the aim when basing text annotation or classification on ChatGPT. 

Traditional human-annotation tasks usually involve the assessments of more than one coder. This approach is connected to hopes that pooling the assessment of multiple coders increases the validity and reliability of the measure, compared to depending on one potentially biased coder only. This concept can be transferred to the ChatGPT context by repeating the same classification task. Since ChatGPT has no memory of past conversations in this study’s setting, repetitions of the same input configuration are independent. Hence, repeating the same prompt several times allows to pool assessments and base the classification outputs on a majority decision. For this binary classification task, this means that the label (i.e., News or not News) with the highest frequency among repetitions is taken as final classification output for the given input. To see if pooling improves consistency, three classification regimes are compared for all consistency comparisons in this study: (i) each input is classified once, no pooling takes place, (ii) each input is classified three times, majority decision on the classification output, (iii) each input is classified ten times, majority decision on the classification output.  

\section*{Results}

Figure \ref{fig2} reports the consistency for three different classification regimes when comparing the classification outputs of identical prompts (n = 2340) for two temperature settings (i.e., 0.25 and 1). Because each identical prompt was repeated ten times for each temperature setting, different classification regimes can be compared. When only considering the first classification output of each prompt for both temperature settings under the first (no pooling) regime (i), the consistency between both sets of output is $\alpha$ = 0.75, failing Krippendorff’s recommendation for reasonable reliabilities \parencite{krippendorff_content_2004}. This consistency improves, when pooling the output of more repetitions of the same prompt and choosing the classification with the highest frequency. Hence, when pooling classification outputs of ten repetitions and taking the majority output for each temperature setting, consistency increases to $\alpha$ = 0.91. 

\begin{figure}[h]
\centering
\caption{Comparing output consistency for two temperature settings}
{\sffamily
\begin{tikzpicture}[x=1pt,y=1pt]
\definecolor{fillColor}{RGB}{255,255,255}
\path[use as bounding box,fill=fillColor,fill opacity=0.00] (0,0) rectangle (144.54,180.67);
\begin{scope}
\path[clip] ( 34.16, 30.69) rectangle (139.04,175.17);
\definecolor{drawColor}{gray}{0.92}

\path[draw=drawColor,line width= 0.3pt,line join=round] ( 34.16, 53.67) --
	(139.04, 53.67);

\path[draw=drawColor,line width= 0.3pt,line join=round] ( 34.16, 86.51) --
	(139.04, 86.51);

\path[draw=drawColor,line width= 0.3pt,line join=round] ( 34.16,119.35) --
	(139.04,119.35);

\path[draw=drawColor,line width= 0.3pt,line join=round] ( 34.16,152.19) --
	(139.04,152.19);

\path[draw=drawColor,line width= 0.6pt,line join=round] ( 34.16, 37.25) --
	(139.04, 37.25);

\path[draw=drawColor,line width= 0.6pt,line join=round] ( 34.16, 70.09) --
	(139.04, 70.09);

\path[draw=drawColor,line width= 0.6pt,line join=round] ( 34.16,102.93) --
	(139.04,102.93);

\path[draw=drawColor,line width= 0.6pt,line join=round] ( 34.16,135.77) --
	(139.04,135.77);

\path[draw=drawColor,line width= 0.6pt,line join=round] ( 34.16,168.61) --
	(139.04,168.61);
\definecolor{drawColor}{RGB}{0,0,255}
\definecolor{fillColor}{RGB}{0,0,255}

\path[draw=drawColor,line width= 0.4pt,line join=round,line cap=round,fill=fillColor] ( 53.82, 87.82) circle (  1.43);

\path[draw=drawColor,line width= 0.4pt,line join=round,line cap=round,fill=fillColor] ( 86.60,114.42) circle (  1.43);

\path[draw=drawColor,line width= 0.4pt,line join=round,line cap=round,fill=fillColor] (119.37,137.41) circle (  1.43);

\path[draw=drawColor,line width= 0.6pt,line join=round] ( 51.53, 98.66) --
	( 56.12, 98.66);

\path[draw=drawColor,line width= 0.6pt,line join=round] ( 53.82, 98.66) --
	( 53.82, 77.32);

\path[draw=drawColor,line width= 0.6pt,line join=round] ( 51.53, 77.32) --
	( 56.12, 77.32);

\path[draw=drawColor,line width= 0.6pt,line join=round] ( 84.30,123.29) --
	( 88.89,123.29);

\path[draw=drawColor,line width= 0.6pt,line join=round] ( 86.60,123.29) --
	( 86.60,105.56);

\path[draw=drawColor,line width= 0.6pt,line join=round] ( 84.30,105.56) --
	( 88.89,105.56);

\path[draw=drawColor,line width= 0.6pt,line join=round] (117.08,144.31) --
	(121.67,144.31);

\path[draw=drawColor,line width= 0.6pt,line join=round] (119.37,144.31) --
	(119.37,130.19);

\path[draw=drawColor,line width= 0.6pt,line join=round] (117.08,130.19) --
	(121.67,130.19);
\end{scope}
\begin{scope}
\path[clip] (  0.00,  0.00) rectangle (144.54,180.67);
\definecolor{drawColor}{gray}{0.30}

\node[text=drawColor,anchor=base east,inner sep=0pt, outer sep=0pt, scale=  0.88] at ( 29.21, 34.22) {0.6};

\node[text=drawColor,anchor=base east,inner sep=0pt, outer sep=0pt, scale=  0.88] at ( 29.21, 67.06) {0.7};

\node[text=drawColor,anchor=base east,inner sep=0pt, outer sep=0pt, scale=  0.88] at ( 29.21, 99.90) {0.8};

\node[text=drawColor,anchor=base east,inner sep=0pt, outer sep=0pt, scale=  0.88] at ( 29.21,132.74) {0.9};

\node[text=drawColor,anchor=base east,inner sep=0pt, outer sep=0pt, scale=  0.88] at ( 29.21,165.58) {1.0};
\end{scope}
\begin{scope}
\path[clip] (  0.00,  0.00) rectangle (144.54,180.67);
\definecolor{drawColor}{gray}{0.30}

\node[text=drawColor,anchor=base,inner sep=0pt, outer sep=0pt, scale=  0.88] at ( 53.82, 19.68) {i};

\node[text=drawColor,anchor=base,inner sep=0pt, outer sep=0pt, scale=  0.88] at ( 86.60, 19.68) {ii};

\node[text=drawColor,anchor=base,inner sep=0pt, outer sep=0pt, scale=  0.88] at (119.37, 19.68) {iii};
\end{scope}
\begin{scope}
\path[clip] (  0.00,  0.00) rectangle (144.54,180.67);
\definecolor{drawColor}{RGB}{0,0,0}

\node[text=drawColor,anchor=base,inner sep=0pt, outer sep=0pt, scale=  1.10] at ( 86.60,  6.64) {Classification regime};
\end{scope}
\begin{scope}
\path[clip] (  0.00,  0.00) rectangle (144.54,180.67);
\definecolor{drawColor}{RGB}{0,0,0}

\node[text=drawColor,rotate= 90.00,anchor=base,inner sep=0pt, outer sep=0pt, scale=  1.10] at ( 09,102.93) {Krippendorff's Alpha};
\end{scope}
\end{tikzpicture}
} 
  \captionsetup{font=footnotesize, skip=2pt} 
  \caption*{Note: Pairwise comparison for n = 468, error bars give 95\% confidence interval}
  \label{fig2}

\end{figure}

Figure \ref{fig3} presents the classification consistency for three classification regimes when comparing the classification outputs for two different instructions, holding everything else constant (n = 468). The instructions, for which the classification output is compared, are identical except that one uses the wording “classify” and “classification”, while the other uses “rate” and “rating”. For all classification regimes, Krippendorff’s Alpha for this comparison does not exceed 0.6. Again, consistency increases when pooling the classification output of several repetitions. Pairwise comparison of all 45 combinations of the ten instructions does by and large not change these impressions (min $\alpha$ = 0.24, mean $\alpha$ = 0.43, max $\alpha$ = 0.7 for classification regime (i)). 

\begin{figure}[t]
\centering
\caption{Comparing output consistency for two different instructions}
{\sffamily
\begin{tikzpicture}[x=1pt,y=1pt]
\definecolor{fillColor}{RGB}{255,255,255}
\path[use as bounding box,fill=fillColor,fill opacity=0.00] (0,0) rectangle (144.54,180.67);
\begin{scope}
\path[clip] ( 34.16, 30.69) rectangle (139.04,175.17);
\definecolor{drawColor}{gray}{0.92}

\path[draw=drawColor,line width= 0.3pt,line join=round] ( 34.16, 46.64) --
	(139.04, 46.64);

\path[draw=drawColor,line width= 0.3pt,line join=round] ( 34.16, 65.40) --
	(139.04, 65.40);

\path[draw=drawColor,line width= 0.3pt,line join=round] ( 34.16, 84.17) --
	(139.04, 84.17);

\path[draw=drawColor,line width= 0.3pt,line join=round] ( 34.16,102.93) --
	(139.04,102.93);

\path[draw=drawColor,line width= 0.3pt,line join=round] ( 34.16,121.70) --
	(139.04,121.70);

\path[draw=drawColor,line width= 0.3pt,line join=round] ( 34.16,140.46) --
	(139.04,140.46);

\path[draw=drawColor,line width= 0.3pt,line join=round] ( 34.16,159.22) --
	(139.04,159.22);

\path[draw=drawColor,line width= 0.6pt,line join=round] ( 34.16, 37.25) --
	(139.04, 37.25);

\path[draw=drawColor,line width= 0.6pt,line join=round] ( 34.16, 56.02) --
	(139.04, 56.02);

\path[draw=drawColor,line width= 0.6pt,line join=round] ( 34.16, 74.78) --
	(139.04, 74.78);

\path[draw=drawColor,line width= 0.6pt,line join=round] ( 34.16, 93.55) --
	(139.04, 93.55);

\path[draw=drawColor,line width= 0.6pt,line join=round] ( 34.16,112.31) --
	(139.04,112.31);

\path[draw=drawColor,line width= 0.6pt,line join=round] ( 34.16,131.08) --
	(139.04,131.08);

\path[draw=drawColor,line width= 0.6pt,line join=round] ( 34.16,149.84) --
	(139.04,149.84);

\path[draw=drawColor,line width= 0.6pt,line join=round] ( 34.16,168.61) --
	(139.04,168.61);
\definecolor{drawColor}{RGB}{0,0,255}
\definecolor{fillColor}{RGB}{0,0,255}

\path[draw=drawColor,line width= 0.4pt,line join=round,line cap=round,fill=fillColor] ( 53.82, 76.28) circle (  1.43);

\path[draw=drawColor,line width= 0.4pt,line join=round,line cap=round,fill=fillColor] ( 86.60, 82.66) circle (  1.43);

\path[draw=drawColor,line width= 0.4pt,line join=round,line cap=round,fill=fillColor] (119.37, 94.11) circle (  1.43);

\path[draw=drawColor,line width= 0.6pt,line join=round] ( 52.67, 97.30) --
	( 54.97, 97.30);

\path[draw=drawColor,line width= 0.6pt,line join=round] ( 53.82, 97.30) --
	( 53.82, 55.64);

\path[draw=drawColor,line width= 0.6pt,line join=round] ( 52.67, 55.64) --
	( 54.97, 55.64);

\path[draw=drawColor,line width= 0.6pt,line join=round] ( 85.45,105.18) --
	( 87.75,105.18);

\path[draw=drawColor,line width= 0.6pt,line join=round] ( 86.60,105.18) --
	( 86.60, 61.08);

\path[draw=drawColor,line width= 0.6pt,line join=round] ( 85.45, 61.08) --
	( 87.75, 61.08);

\path[draw=drawColor,line width= 0.6pt,line join=round] (118.23,114.19) --
	(120.52,114.19);

\path[draw=drawColor,line width= 0.6pt,line join=round] (119.37,114.19) --
	(119.37, 74.97);

\path[draw=drawColor,line width= 0.6pt,line join=round] (118.23, 74.97) --
	(120.52, 74.97);
\end{scope}
\begin{scope}
\path[clip] (  0.00,  0.00) rectangle (144.54,180.67);
\definecolor{drawColor}{gray}{0.30}

\node[text=drawColor,anchor=base east,inner sep=0pt, outer sep=0pt, scale=  0.88] at ( 29.21, 34.22) {0.3};

\node[text=drawColor,anchor=base east,inner sep=0pt, outer sep=0pt, scale=  0.88] at ( 29.21, 52.99) {0.4};

\node[text=drawColor,anchor=base east,inner sep=0pt, outer sep=0pt, scale=  0.88] at ( 29.21, 71.75) {0.5};

\node[text=drawColor,anchor=base east,inner sep=0pt, outer sep=0pt, scale=  0.88] at ( 29.21, 90.52) {0.6};

\node[text=drawColor,anchor=base east,inner sep=0pt, outer sep=0pt, scale=  0.88] at ( 29.21,109.28) {0.7};

\node[text=drawColor,anchor=base east,inner sep=0pt, outer sep=0pt, scale=  0.88] at ( 29.21,128.05) {0.8};

\node[text=drawColor,anchor=base east,inner sep=0pt, outer sep=0pt, scale=  0.88] at ( 29.21,146.81) {0.9};

\node[text=drawColor,anchor=base east,inner sep=0pt, outer sep=0pt, scale=  0.88] at ( 29.21,165.58) {1.0};
\end{scope}
\begin{scope}
\path[clip] (  0.00,  0.00) rectangle (144.54,180.67);
\definecolor{drawColor}{gray}{0.30}

\node[text=drawColor,anchor=base,inner sep=0pt, outer sep=0pt, scale=  0.88] at ( 53.82, 19.68) {i};

\node[text=drawColor,anchor=base,inner sep=0pt, outer sep=0pt, scale=  0.88] at ( 86.60, 19.68) {ii};

\node[text=drawColor,anchor=base,inner sep=0pt, outer sep=0pt, scale=  0.88] at (119.37, 19.68) {iii};
\end{scope}
\begin{scope}
\path[clip] (  0.00,  0.00) rectangle (144.54,180.67);
\definecolor{drawColor}{RGB}{0,0,0}

\node[text=drawColor,anchor=base,inner sep=0pt, outer sep=0pt, scale=  1.10] at ( 86.60,  7.64) {Classification regime};
\end{scope}
\begin{scope}
\path[clip] (  0.00,  0.00) rectangle (144.54,180.67);
\definecolor{drawColor}{RGB}{0,0,0}

\node[text=drawColor,rotate= 90.00,anchor=base,inner sep=0pt, outer sep=0pt, scale=  1.10] at ( 09,102.93) {Krippendorff's Alpha};
\end{scope}
\end{tikzpicture}
} 
  \captionsetup{font=footnotesize, skip=2pt} 
  \caption*{Note: Pairwise comparison for n = 468, error bars give 95\% confidence interval}
  \label{fig3}

\end{figure}

Finally, figure \ref{fig4} reports the classification consistency for three classification regimes when comparing the classification outputs of repetitions of identical prompts (i.e., everything is held constant, n = 2340). Results are split for both temperature settings. It becomes apparent that within-consistency is higher for the lower temperature setting, with Krippen- dorff’s Alpha for all three classification regimes above 0.9. This indicates a reasonable reliability. In contrast, for the higher temperature settings, only within-consistency for the highest classification regime reaches a reasonable reliability of $\alpha$ = 0.85. As this study took ten repetitions per input configuration, a comparison for the usual classification regime (iii) is not available for this analysis as this would have required at least 20 repetitions (i.e., comparing the pooled output of ten repetitions vs. the pooled output of ten other repetitions). So, here the pooled output of five repetitions each were compared.

\vspace{0.2cm}
\begin{figure}[h]
\centering
\caption{Comparing output consistency for repetitions of identical inputs}
{\sffamily
\begin{tikzpicture}[x=1pt,y=1pt]
\definecolor{fillColor}{RGB}{255,255,255}
\path[use as bounding box,fill=fillColor,fill opacity=0.00] (0,0) rectangle (227.65,180.67);
\begin{scope}
\path[clip] ( 34.16, 30.69) rectangle (138.38,175.17);
\definecolor{drawColor}{gray}{0.92}

\path[draw=drawColor,line width= 0.3pt,line join=round] ( 34.16, 53.67) --
	(138.38, 53.67);

\path[draw=drawColor,line width= 0.3pt,line join=round] ( 34.16, 86.51) --
	(138.38, 86.51);

\path[draw=drawColor,line width= 0.3pt,line join=round] ( 34.16,119.35) --
	(138.38,119.35);

\path[draw=drawColor,line width= 0.3pt,line join=round] ( 34.16,152.19) --
	(138.38,152.19);

\path[draw=drawColor,line width= 0.6pt,line join=round] ( 34.16, 37.25) --
	(138.38, 37.25);

\path[draw=drawColor,line width= 0.6pt,line join=round] ( 34.16, 70.09) --
	(138.38, 70.09);

\path[draw=drawColor,line width= 0.6pt,line join=round] ( 34.16,102.93) --
	(138.38,102.93);

\path[draw=drawColor,line width= 0.6pt,line join=round] ( 34.16,135.77) --
	(138.38,135.77);

\path[draw=drawColor,line width= 0.6pt,line join=round] ( 34.16,168.61) --
	(138.38,168.61);
\definecolor{drawColor}{RGB}{0,0,255}
\definecolor{fillColor}{RGB}{0,0,255}

\path[draw=drawColor,line width= 0.4pt,line join=round,line cap=round,fill=fillColor] ( 53.70,136.75) circle (  1.43);

\path[draw=drawColor,line width= 0.4pt,line join=round,line cap=round,fill=fillColor] ( 86.27,147.26) circle (  1.43);

\path[draw=drawColor,line width= 0.4pt,line join=round,line cap=round,fill=fillColor] (118.84,151.20) circle (  1.43);
\definecolor{drawColor}{RGB}{255,0,0}
\definecolor{fillColor}{RGB}{255,0,0}

\path[draw=drawColor,line width= 0.4pt,line join=round,line cap=round,fill=fillColor] ( 53.70, 63.85) circle (  1.43);

\path[draw=drawColor,line width= 0.4pt,line join=round,line cap=round,fill=fillColor] ( 86.27,102.93) circle (  1.43);

\path[draw=drawColor,line width= 0.4pt,line join=round,line cap=round,fill=fillColor] (118.84,119.02) circle (  1.43);
\definecolor{drawColor}{RGB}{0,0,255}

\path[draw=drawColor,line width= 0.6pt,line join=round] ( 51.42,143.65) --
	( 55.98,143.65);

\path[draw=drawColor,line width= 0.6pt,line join=round] ( 53.70,143.65) --
	( 53.70,129.86);

\path[draw=drawColor,line width= 0.6pt,line join=round] ( 51.42,129.86) --
	( 55.98,129.86);

\path[draw=drawColor,line width= 0.6pt,line join=round] ( 83.99,153.17) --
	( 88.55,153.17);

\path[draw=drawColor,line width= 0.6pt,line join=round] ( 86.27,153.17) --
	( 86.27,141.35);

\path[draw=drawColor,line width= 0.6pt,line join=round] ( 83.99,141.35) --
	( 88.55,141.35);

\path[draw=drawColor,line width= 0.6pt,line join=round] (116.56,156.46) --
	(121.12,156.46);

\path[draw=drawColor,line width= 0.6pt,line join=round] (118.84,156.46) --
	(118.84,145.95);

\path[draw=drawColor,line width= 0.6pt,line join=round] (116.56,145.95) --
	(121.12,145.95);
\definecolor{drawColor}{RGB}{255,0,0}

\path[draw=drawColor,line width= 0.6pt,line join=round] ( 51.42, 75.67) --
	( 55.98, 75.67);

\path[draw=drawColor,line width= 0.6pt,line join=round] ( 53.70, 75.67) --
	( 53.70, 51.70);

\path[draw=drawColor,line width= 0.6pt,line join=round] ( 51.42, 51.70) --
	( 55.98, 51.70);

\path[draw=drawColor,line width= 0.6pt,line join=round] ( 83.99,112.78) --
	( 88.55,112.78);

\path[draw=drawColor,line width= 0.6pt,line join=round] ( 86.27,112.78) --
	( 86.27, 93.08);

\path[draw=drawColor,line width= 0.6pt,line join=round] ( 83.99, 93.08) --
	( 88.55, 93.08);

\path[draw=drawColor,line width= 0.6pt,line join=round] (116.56,127.56) --
	(121.12,127.56);

\path[draw=drawColor,line width= 0.6pt,line join=round] (118.84,127.56) --
	(118.84,110.15);

\path[draw=drawColor,line width= 0.6pt,line join=round] (116.56,110.15) --
	(121.12,110.15);
\end{scope}
\begin{scope}
\path[clip] (  0.00,  0.00) rectangle (227.65,180.67);
\definecolor{drawColor}{gray}{0.30}

\node[text=drawColor,anchor=base east,inner sep=0pt, outer sep=0pt, scale=  0.88] at ( 29.21, 34.22) {0.6};

\node[text=drawColor,anchor=base east,inner sep=0pt, outer sep=0pt, scale=  0.88] at ( 29.21, 67.06) {0.7};

\node[text=drawColor,anchor=base east,inner sep=0pt, outer sep=0pt, scale=  0.88] at ( 29.21, 99.90) {0.8};

\node[text=drawColor,anchor=base east,inner sep=0pt, outer sep=0pt, scale=  0.88] at ( 29.21,132.74) {0.9};

\node[text=drawColor,anchor=base east,inner sep=0pt, outer sep=0pt, scale=  0.88] at ( 29.21,165.58) {1.0};
\end{scope}
\begin{scope}
\path[clip] (  0.00,  0.00) rectangle (227.65,180.67);
\definecolor{drawColor}{gray}{0.30}

\node[text=drawColor,anchor=base,inner sep=0pt, outer sep=0pt, scale=  0.88] at ( 53.70, 19.68) {i};

\node[text=drawColor,anchor=base,inner sep=0pt, outer sep=0pt, scale=  0.88] at ( 86.27, 19.68) {ii};

\node[text=drawColor,anchor=base,inner sep=0pt, outer sep=0pt, scale=  0.88] at (118.84, 19.68) {5*};
\end{scope}
\begin{scope}
\path[clip] (  0.00,  0.00) rectangle (227.65,180.67);
\definecolor{drawColor}{RGB}{0,0,0}

\node[text=drawColor,anchor=base,inner sep=0pt, outer sep=0pt, scale=  1.10] at ( 86.27,  7.64) {Classification regime};
\end{scope}
\begin{scope}
\path[clip] (  0.00,  0.00) rectangle (227.65,180.67);
\definecolor{drawColor}{RGB}{0,0,0}

\node[text=drawColor,rotate= 90.00,anchor=base,inner sep=0pt, outer sep=0pt, scale=  1.10] at ( 09,102.93) {Krippendorff's Alpha};
\end{scope}
\begin{scope}
\path[clip] (  0.00,  0.00) rectangle (227.65,180.67);
\definecolor{drawColor}{RGB}{0,0,0}

\node[text=drawColor,anchor=base west,inner sep=0pt, outer sep=0pt, scale=  1.10] at (154.88,116.35) {Temperature};
\end{scope}
\begin{scope}
\path[clip] (  0.00,  0.00) rectangle (227.65,180.67);
\definecolor{drawColor}{RGB}{0,0,255}
\definecolor{fillColor}{RGB}{0,0,255}

\path[draw=drawColor,line width= 0.4pt,line join=round,line cap=round,fill=fillColor] (162.11,102.55) circle (  1.43);
\end{scope}
\begin{scope}
\path[clip] (  0.00,  0.00) rectangle (227.65,180.67);
\definecolor{drawColor}{RGB}{0,0,255}

\path[draw=drawColor,line width= 0.6pt,line join=round] (156.33,102.55) -- (167.89,102.55);
\end{scope}
\begin{scope}
\path[clip] (  0.00,  0.00) rectangle (227.65,180.67);
\definecolor{drawColor}{RGB}{255,0,0}
\definecolor{fillColor}{RGB}{255,0,0}

\path[draw=drawColor,line width= 0.4pt,line join=round,line cap=round,fill=fillColor] (162.11, 88.10) circle (  1.43);
\end{scope}
\begin{scope}
\path[clip] (  0.00,  0.00) rectangle (227.65,180.67);
\definecolor{drawColor}{RGB}{255,0,0}

\path[draw=drawColor,line width= 0.6pt,line join=round] (156.33, 88.10) -- (167.89, 88.10);
\end{scope}
\begin{scope}
\path[clip] (  0.00,  0.00) rectangle (227.65,180.67);
\definecolor{drawColor}{RGB}{0,0,0}

\node[text=drawColor,anchor=base west,inner sep=0pt, outer sep=0pt, scale=  0.88] at (174.84, 99.52) {0.25};
\end{scope}
\begin{scope}
\path[clip] (  0.00,  0.00) rectangle (227.65,180.67);
\definecolor{drawColor}{RGB}{0,0,0}

\node[text=drawColor,anchor=base west,inner sep=0pt, outer sep=0pt, scale=  0.88] at (174.84, 85.07) {1};
\end{scope}
\end{tikzpicture}
} 
  \captionsetup{font=footnotesize, skip=6pt} 
  \begin{minipage}{0.84\textwidth}
  \vspace{0.2cm}
  \caption*{Note: Pairwise comparison for n = 2340, error bars give 95\% confidence interval. In this setting, the third classification regime pools the outcome of 5 instead of 10 repetitions}
  \end{minipage}
  \label{fig4}

\end{figure} 
 
\section*{Discussion}

Results of the comparison between two temperature settings show that the classification output by two different temperature settings is not consistent. This is not surprising as ChatGPT’s outputs from a higher temperature setting are meant to be more random \parencite{openaia}. Accordingly, the classifications between both settings differ. In parallel, consistency within the same temperature setting is higher for lower temperature settings as demonstrated in Figure \ref{fig3}. Furthermore, it is likely that the length of prompts and complexity of the classification task interact with the randomness of the temperature settings. However, how temperature settings impact the validity will need to be evaluated for each study (e.g., comparing ChatGPT’s classification outputs with reference classifications). For high temperature settings, the comparison should be based on a larger set of classifications to prevent misleading performance assessments caused by ChatGPT’s randomness. 
Second, the prompt is decisive for the output and even small variations can cause large inconsistencies. For a human coder, the change of a single word in the instructions would likely not have caused large inconsistencies as the overall meaning was barely changed. When aiming to use ChatGPT for zero-shot text annotation and classification this has implications for the preparation of instructions (i.e., codebook). The findings of this study indicate that the same codebook, which works well for human coders (who might be less sensitive to small changes in instructions), does not necessarily perform well for ChatGPT. This can pose an arduous problem as slight variations could largely impact outputs and their validity and reliability. Future studies could investigate efficient prompting strategies for text annotation contexts \parencite{argyle_out_2023}.  

While the other factors can be optimized and held constant, inconsistencies within repetitions point to randomness in ChatGPT that affects all classification outcomes, making this the most important source for possible inconsistencies. Results show that within-repetition inconsistencies are starkly influenced by the temperature setting, which steers the randomness of the output. When randomness is reduced due to low temperature settings, output consistency for identical input is high. When temperature settings are high, consistency is comparatively low. Since consistency and reliability do not necessarily relate to validity, it is possible that in some contexts high temperature settings are preferable, even though reliability is reduced. In such cases, pooling of the classification outputs of repetitions of the same input can improve consistency (but still requires validation). 

For all three factors investigated in this study, majority decisions on repeated classification outputs of the same input led to improvements in consistency. This is a valuable strategy to protect classification results against the randomness of ChatGPT. Therefore, building a majority decision based on three, ideally more repetitions of each single input is recommended when using ChatGPT for zero-shot text annotation. This also depends on the resources as repetitions increase computing time and costs, making ChatGPT potentially less attractive \parencite{gilardi_chatgpt_2023}.\footnote{Currently (April 2023), OpenAI has a default cap on its API usage per month which is at \$120. The cost for this study with around 48,000 inputs (76 million tokens) was \$152.} The inclusion of ChatGPT in text annotation and analysis software must also be viewed critically in this respect as users might falsely assume reliable outcomes by ChatGPT while outputs are likely not pooled and hence not reliable. And even for pooled outputs, validation should always be a priority as this might increase reliability but does necessarily relate to or inform about the validity. 

The texts used for this study are comparatively noisy (as a lot of website artefacts were included) and in German. It is possible that consistency is higher for English texts, shorter texts, and texts with a clearer structure. However, thorough validation is always necessary as the complexity of the task and the validity and reliability of ChatGPT on particular tasks is not known in advance. Furthermore, although the data and classification task are based on a real-world problem \parencite{reiss_dissecting_2023}, the instructions were slightly adjusted to accommodate for the ChatGPT context. This does not impinge the consistency analysis in this study, but the classification situation is not comparable to the human labeling which is why no comparison with the human labels are made. 

This study is based on gpt-3.5-turbo, which is marketed as the “most capable and cost effective model in the GPT-3.5 family” \parencite{openaib} and which is the basis of the popular ChatGPT web application \parencite{brockman_introducing_2023}. Future studies should also test gpt-4, which is again more capable than any gpt-3.5 model but comes with more than 10 times the costs (\$0.03/1K tokens vs. \$0.002/1K tokens). At the time of writing this study, API access to gpt-4 is still limited and the author had no access to it. A very limited test via the web interface of some prompts that had the least consistency for gpt-3.5 showed much improved, although not perfect consistency when using gpt-4. 

\textbf{Conclusion}

This study assessed the consistency of ChatGPT’s zero-shot capabilities for text annotation. Several factors contribute to the fact that the overall consistency of zero-shot text annotation by ChatGPT can be considerably below common scientific standards. First, lower temperature settings make annotations by ChatGPT more deterministic and hence consistent, but this could interact with the validity of outputs as high temperature settings could provide more valid but less reliable output. Second, the exact prompt is decisive and consistency between two prompts can be low, also for minuscule wording alterations between two prompts. Third, even repeating the identical input configurations (i.e., same prompt, same parameter settings) can lead to different outputs, questioning the reliability of ChatGPT for text annotation and classification. Furthermore, the extent of inconsistencies and relation to the validity of the classification likely varies depending on the classification or annotation task. All in all, even though pooling the outputs of repetitions of the same input can increase reliability, findings of this study suggest that ChatGPT for zero-shot text annotation should only be used with caution: robust validation of ChatGPT’s output is strongly recommended. While some promising studies suggest the usability of ChatGPT for text annotation and classification, it is important to note that this can only be guaranteed when validation is included (e.g., comparing against human-annotated references). The unsupervised use should be avoided. 

\vfill
\section*{Acknowledgements}

I want to thank Michael Jacobs, Kiran Kappeler, Valerie Hase, Aleksandra Urman, Manuel Issel, Nico Pfiffner, and Fjona Gerber for helpful comments and valuable input at various stages of this project and my division led by Michael Latzer for the opportunity to use the data.
 
\section*{Funding}

Parts of this work were supported by the Swiss National Science Foundation [176443].

\pagebreak
\section*{References}
\printbibliography[heading = none]

\end{document}